\documentclass[lettersize,journal]{IEEEtran}
\usepackage{amsmath,amsfonts}
\usepackage{algpseudocode}
\usepackage{array}
\usepackage{subfigure}
\usepackage{textcomp}
\usepackage{stfloats}
\usepackage{url}
\usepackage{verbatim}
\usepackage{graphicx}
\usepackage{cite}
\usepackage{color}

\usepackage{hyperref}
\hypersetup{
colorlinks=true,
linkcolor=blue,
filecolor=blue,
urlcolor=red,
citecolor=green,
}

\usepackage[ruled,vlined,linesnumbered]{algorithm2e}

\newtheorem{problem}{\bf Problem}[section]
\newtheorem{definition}{\bf Definition}[section]

\begin{document}

\title{NuRF: Nudging the Particle Filter in Radiance Fields for Robot Visual Localization}

\author{Wugang Meng, Tianfu Wu, Huan Yin and Fumin Zhang
\thanks{All authors are with the Department of Electronic and Computer Engineering, Hong Kong University of Science and Technology, Hong Kong SAR. The work described in this paper was supported by grants AoE/E-601/24-N, 16203223, and C6029-23G from the Research Grants Council of the Hong Kong Special Administrative Region, China. \textit{(Corresponding author: Huan Yin)}
}
}

\markboth{Accepted for Publication in IEEE Transactions on Cognitive and Developmental Systems}%
{Shell \MakeLowercase{\textit{et al.}}: A Sample Article Using IEEEtran.cls for IEEE Journals}


\maketitle

\begin{abstract}
Can we localize a robot on a map only using monocular vision? This study presents NuRF, an adaptive and nudged particle filter framework in radiance fields for 6-DoF robot visual localization. NuRF leverages recent advancements in radiance fields and visual place recognition. Conventional visual place recognition meets the challenges of data sparsity and artifact-induced inaccuracies. By utilizing radiance field-generated novel views, NuRF enhances visual localization performance and combines coarse global localization with the fine-grained pose tracking of a particle filter, ensuring continuous and precise localization. Experimentally, our method converges 7 times faster than existing Monte Carlo-based methods and achieves localization accuracy within 1 meter, offering an efficient and resilient solution for indoor visual localization.
\end{abstract}


\section{Introduction}

\IEEEPARstart{V}{isual} localization on a map is a critical topic for robot navigation, which is close to how humans perceive the world. An ideal visual localization system is desired to accurately localize robots not only globally but also continuously~\cite{burgard1998integrating}.
Existing feature-based techniques, like perspective-n-point (PnP) \cite{fischler1981random} offer stable and accurate continuous pose tracking for robots, they struggle to directly address global localization in cases of localization failure \cite{yin2024survey}. On the other hand, learning-based approaches like visual place recognition (VPR) \cite{lowry2015visual,schubert2023visual,miao2024survey} can regress or retrieve poses through trained networks from scratch. These approaches fall short of providing a comprehensive and cost-effective solution for continuous 6-degree-of-freedom (6-DoF) pose tracking.
There is a growing need for a unified framework that integrates global localization and pose tracking for robot visual localization, enabling the use of one localization method on one map for two tasks. 

Recent advancements in radiance field rendering, particularly neural radiance fields (NeRF)\cite{mildenhall2021nerf} and 3D Gaussian splatting (3DGS)\cite{kerbl20233d}, have shown significant potential in overcoming the limitations of traditional visual localization. Radiance field-based maps can enhance visual-based localization by providing dense, high-quality, and renderable 3D reconstructions of environments~\cite{liu2023nerfloc,maggio2023loc,sun2023icomma,jiang20243dgs}, thereby improving localization capabilities in texture-less or occluded regions. Recent studies~\cite{maggio2023loc,sun2023icomma} demonstrate the feasibility of utilizing radiance fields for visual global localization. However, casually captured radiance fields often suffer from unreal artifacts, such as floaters, which can emerge and are not part of the real view~\cite{warburg2023nerfbusters}. These ghostly artifacts critically impact visual localization in terms of robustness and accuracy. Existing regularizers~\cite{lin2021barf,martin2021nerf,sabour2023robustnerf} are ineffective in detecting or eliminating floaters; alternative optimization methods based on 3D diffusion models~\cite{warburg2023nerfbusters} are too time-consuming for real-time robot localization.


\begin{figure}[t]
    \centering
    \includegraphics[width=9cm]{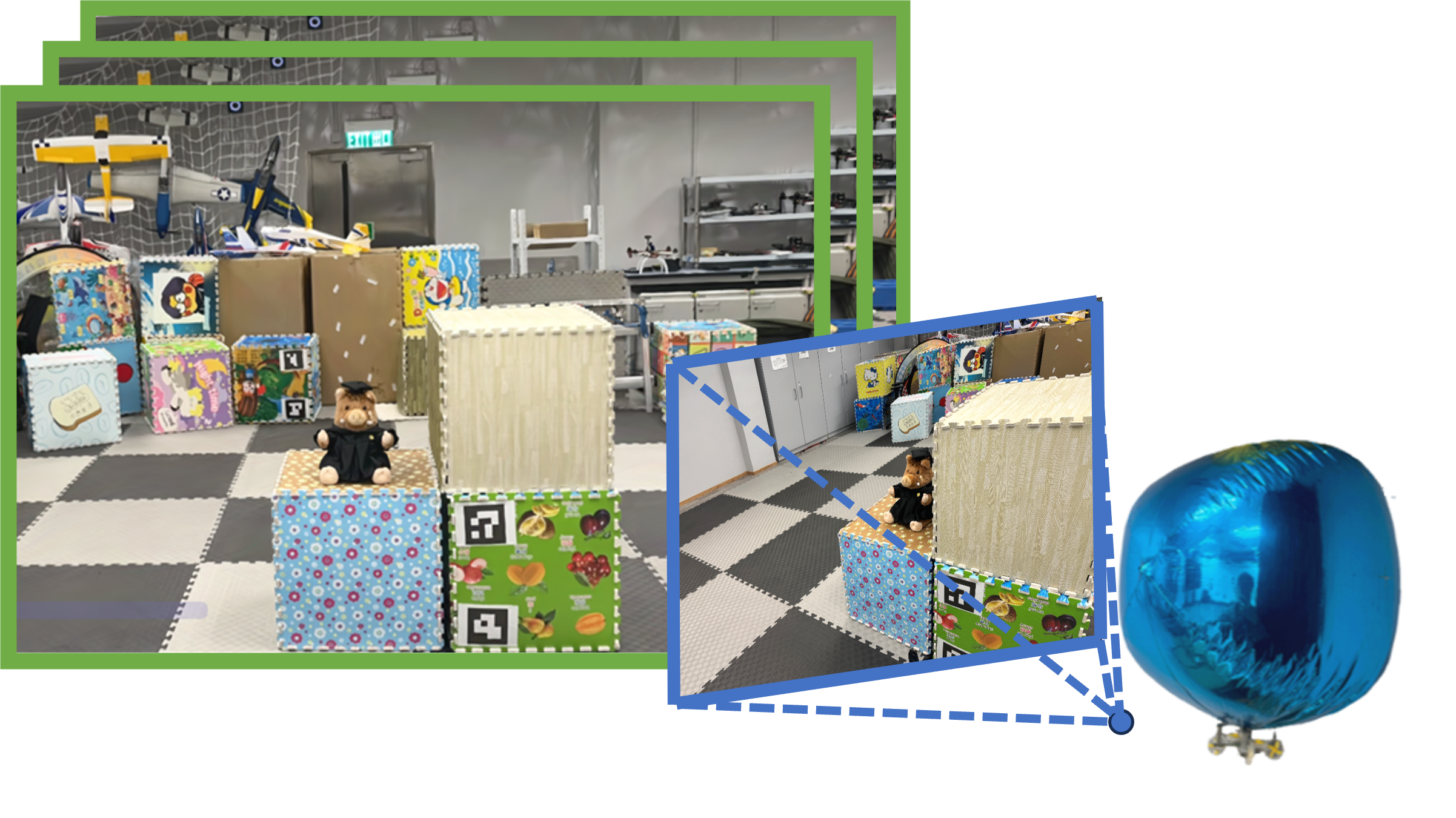}
    \caption{NuRF achieves monocular localization on images generated from radiance fields. The image within the blue box is observed by our blimp robot, while the image in the green box is a reference image rendered from radiance fields..}
    \label{pig_mix}
\end{figure}


To address the limitations above, we propose an adaptive nudged particle filter in radiance field (NuRF). NuRF enhances VPR through radiance field representations and nudges VPR information to mitigate the impact of floaters. First, we utilize the radiance field to generate novel views from new perspectives, overcoming the challenge of sparseness for VPR. 
Second, inspired by a previous work~\cite{lin2024particle}, the nudged particle filter fuses the coarse global localization information from VPR with the particle filter’s fine-grained pose estimation, enabling continuous, global and efficient localization. Third, this hybrid approach also ensures a smooth transition from global localization to accurate pose tracking via an adaptive mechanism. This approach effectively balances computational efficiency and precision, addressing the challenges of robot localization using NeRF as maps. Overall, the contributions are summarized as follows:
\begin{itemize}
    \item We achieve a monocular robot localization by using radiance fields as maps and leveraging VPR techniques.
    \item We propose nudging the particle filter to enhance visual localization performance and overcome the challenge of artifacts in radiance fields.
    \item We design an adaptive scheme that switches between global localization and position tracking for robust and efficient visual localization.
    \item All experiments are conducted with our blimp robot in the real world, with ablated tests.
\end{itemize}

The rest of our paper is organized as follows: Section~\ref{sec:relatedwork} introduces the related work on radiance field, visual place recognition, and nudged particle filter; Section~\ref{sec:problem} presents the problem formulation. We detail the proposed method in Section~\ref{sec:meth} and demonstrate its effectiveness with real-world experiments in Section~\ref{sec:exp}. The conclusions and future studies are summarized in Section~\ref{sec:conclusions}.









\section{Related Work}
\label{sec:relatedwork}

\subsection{Localization in Radiance Fields}
A radiance field is a representation of how light is distributed in three-dimensional space, capturing the interactions between light and surfaces in the environment~\cite{mildenhall2021nerf}. 
It can be seen as a low-dimensional sub-manifold that encompasses all the rendered images of the scene in a high-dimensional pixel space. In recent years, radiance field is a new wave for robotic applications, as summarized in the survey paper~\cite{wang2024nerf}.

Loc-NeRF by Maggio \textit{et al.} \cite{maggio2023loc} introduced a Monte Carlo localization method for estimating the posterior probability of individual camera poses in a given space by rendering multiple random images within a radiance field. These rendered images are then compared to the target image in pixel space, evaluating the pixel-space distances. The recent advancement in explicit neural radiance field Gaussian splatting \cite{kerbl20233d} has achieved a rendering rate of 160 frames per second (fps) per image while requiring less than 500MB of storage \cite{hamdi2024ges}. Moreover, this technology can be implemented using handheld devices such as smartphones \cite{keetha2023splatam}. In the 3DGS-ReLoc by Jiang \textit{et al.}~\cite{jiang20243dgs}, Gaussian Splatting representation was utilized for visual re-localization in urban scenarios, following a coarse-to-fine manner for global localization. Specifically, the camera was first located by similarity comparisons, and then refined by the PnP technique. This indicates that the proposed method might not be practically feasible for engineering implementation.


\subsection{Visual Place Recognition}

VPR is a fundamental problem in robotics and computer vision \cite{lowry2015visual}, aiming to provide a coarse pose estimation by retrieving geo-referenced frames in close proximity. Modern learning-based VPR solutions, such as AnyLoc \cite{keetha2023anyloc}, offer comprehensive and accurate frame retrieval due to their robust representation of both images and maps. 

However, VPR achieves localization by considering the reference pose approximating the query pose. This inherent approximation limits the localization accuracy, especially when there is a significant deviation between the current pose and the queried poses in the database. Consequently, VPR is typically employed as a coarse or initial step in high-precision localization systems \cite{sarlin2019coarse}, which is also analyzed in the survey paper for global LiDAR localization~\cite{yin2024survey}. On the other hand, constructing a database of landmark anchors for reference purposes can be a time-consuming and labor-intensive process. It involves collecting a reference image database from the robot's onboard camera and odometer data as it navigates through the environment.

\subsection{Nudged Particle Filter}
Nudging refers to a sampling approach in particle filtering that guides particles towards regions of high expected likelihood \cite{akyildiz2020nudging}. This technique has been demonstrated to effectively address issues with Bootstrap particle filters~\cite{gordon1993novel} when modeling errors are present, and it's particularly effective when the posterior probabilities concentrate in relatively small areas of the state space, making it well-suited for high-resolution observation models, such as images. In the previous work conducted by Lin \textit{et al.} \cite{lin2024particle}, conducted experiments demonstrating that the utilization of nudging particles enabled accurate tracking of a robot's pose, even in cases where the initial distribution was incorrect and had low variance. 

The preliminary version of Nudged Particle Filter was developed with the objective of rectifying erroneous kinetic modeling assumptions through the use of nudging and did not achieve successful global localization using the nudged particle filter in that particular study. We initially proposed the application of Nudged Particle Filter to address deficiencies in observation models like Radiance Field and global localization.


\section{Problem Formulation}\label{sec:problem}
The visual localization problem can be formulated as an estimation of an unknown posterior distribution. For convenience, we will use the special Euclidean group (SE(3)) to present the camera pose or camera extrinsic, let the pose of a camera-mounted robot at time $\tau$ be the same as the extrinsic matrix of the camera $T_{\tau} \in $ SE(3).  From time $\tau-1$ to $\tau$, the robot pose is integrated by relative transformations $H_{\tau} \in $ SE(3), described as follows:
\begin{equation}
    T_{\tau} = T_{\tau-1} H_{\tau} = \begin{bmatrix}
    \boldsymbol{R}_{\tau-1}\boldsymbol{V}_{\tau} & \boldsymbol{R}_{\tau-1}\boldsymbol{u}_{\tau} + \boldsymbol{t}_{\tau-1}\\
    \boldsymbol{0}^T & 1
    \end{bmatrix}
    \label{eq:pose}
\end{equation}
in which $\boldsymbol{t},\boldsymbol{u} \in \mathbb{R}^3 $ are the translation vectors and $\boldsymbol{R},\boldsymbol{V} \in \mathbb{R}^{3\times3}$ are rotation matrices. For two consecutive poses, we assume the sensors equipped on the robot can capture two images $I_{\tau-1}$,  $I_{\tau}$ and the motion $\hat{H}_{\tau}$. Mathematically, both implicit and explicit radiance fields model can be represented as a function $\{L: \text{SE(3)} \mapsto \mathbb{R}^{W \times H}\}$. It provides a mapping of the extrinsic matrix $T_{\tau}$ to a $W \times H$ image $\hat{I}_{\tau}$ that with minimum render loss $\text{LOSS}(\cdot,\cdot)$.

\begin{equation}
    L(T_{\tau}) = \arg \min_{\hat{I}_{\tau}} \text{LOSS}(I_{\tau},\hat{I}_{\tau})
    \label{eq:rendering}
\end{equation}

Then, the posterior probability density function of the robot pose $T_{\tau}$ can be written as:
\begin{equation}
    \begin{aligned}
        p(T_{\tau}|I_{1:\tau},\hat{H}_{2:\tau},L) = \frac{p(I_{1:\tau}|L,T_{\tau})p(T_{\tau}|\hat{H}_{2:\tau})}{p(I_{1:\tau}|L)} 
    \label{eq:poster}
    \end{aligned}
\end{equation}
and the visual-based localization problem can be described as an estimation problem for unknown distribution, described as follows:
\begin{problem}
    Given a sequence of RGB images $\boldsymbol{I} = [I_1, I_2, \dots, I_{\tau}]$, motion sequence $\boldsymbol{\hat{H}} = [\hat{H}_2, \hat{H}_3, \dots, \hat{H}_{\tau}]$,  and the radiance field model $L$, how to approximate the posterior distribution of the robot pose $T{\tau}$ as described in Equation \eqref{eq:poster}.
    \label{prob:localization}
\end{problem}

Problem~\ref{prob:localization} is a widely studied problem in the robotics literature for unknown distribution estimation and has been solved by Sequential Monte Carlo (SMC) commonly~\cite{doucet2009tutorial}.
Let $\varXi_{\tau} = \{\zeta_{\tau,i}\}^{N}_{i=1}$ be the set of particles at time $\tau$, where each particle $\zeta_{\tau,i} = [w_{\tau,i},T_{\tau,i}]$ with weight $w_{\tau,i}$ and pose $T_{\tau,i}$. Consequently, the posterior probability density function \eqref{eq:poster} can be approximated as follows:
\begin{equation}
    f(z; \varXi_{\tau}) = \eta \sum_{i=1}^{N} w_{\tau,i} \delta(z - T_{\tau,i})
    \label{eq:g_distribution}
\end{equation}
in which $\delta$ is the Dirac delta function, and $\eta$ is a normalization constant. The state transition probability $p(T_{\tau}|\hat{H}_{2:\tau})$ is given by Equation~\eqref{eq:pose}, to minimize the diversity between $f(z; \varXi_{\tau})$ and $ p(T_{\tau}|I_{1:\tau},\hat{H}_{2:\tau},L)$, the un-normalized importance weights based on the likelihood should satisfied~\cite{doucet2000sequential}:
\begin{equation}
    w_{\tau,i} = w_{\tau-1,i} p(I_{\tau}|L,T_{\tau,i})
    \label{eq:likeli}
\end{equation}
and the likelihood $p(I_{\tau}|L,T_{\tau,i})$  can be computed directly by the pixel-wised render loss function $\text{LOSS}(\cdot,\cdot)$ in Equation~\eqref{eq:rendering}.


As presented in Equation~\eqref{eq:rendering}, the radiance field rendering function $L$ is optimized via the pixel-wise loss function $\text{LOSS}(\cdot,\cdot)$, indicating that the weights derived from the loss function could be affected by these artifacts. In this paper, we consider it is possible to eliminate unrealistic artifacts by representing images using advanced visual features (descriptors)~\cite{keetha2023anyloc}, thereby retaining only the visual descriptors of the environment. Consequently, a further key focus of this study is how to selectively adjust a small subset of particles to decrease the feature loss based on VPR information during the particle filter update process to enhance estimation accuracy. This consideration leads to the following Problem \ref{prob:nudge}, where the VPR loss is defined by the cosine distance between the visual feature vector of the rendered image and the observed image.
\begin{problem}
    Given a particle set $\varXi$ and $w$ is calculated by pixel-wised loss. The enhancement of the particle set to better approximate the $p(T_{\tau}|I_{1:\tau},\hat{H}_{2:\tau},L)$ involves minimizing the discrepancy between true distribution and the particle-based estimate $f(z; \varXi^{+})$:
    \begin{equation}
      \varXi^{+} = \arg\min_{\varXi^{+}} \parallel  1 - \frac{p(T_{\tau}|I_{1:\tau},\hat{H}_{2:\tau},L)}{f(z; \varXi^{+})} \parallel
    \end{equation}
    \label{prob:nudge}
\end{problem}

\begin{figure*}[t]
  \centering
  \includegraphics[width=18cm]{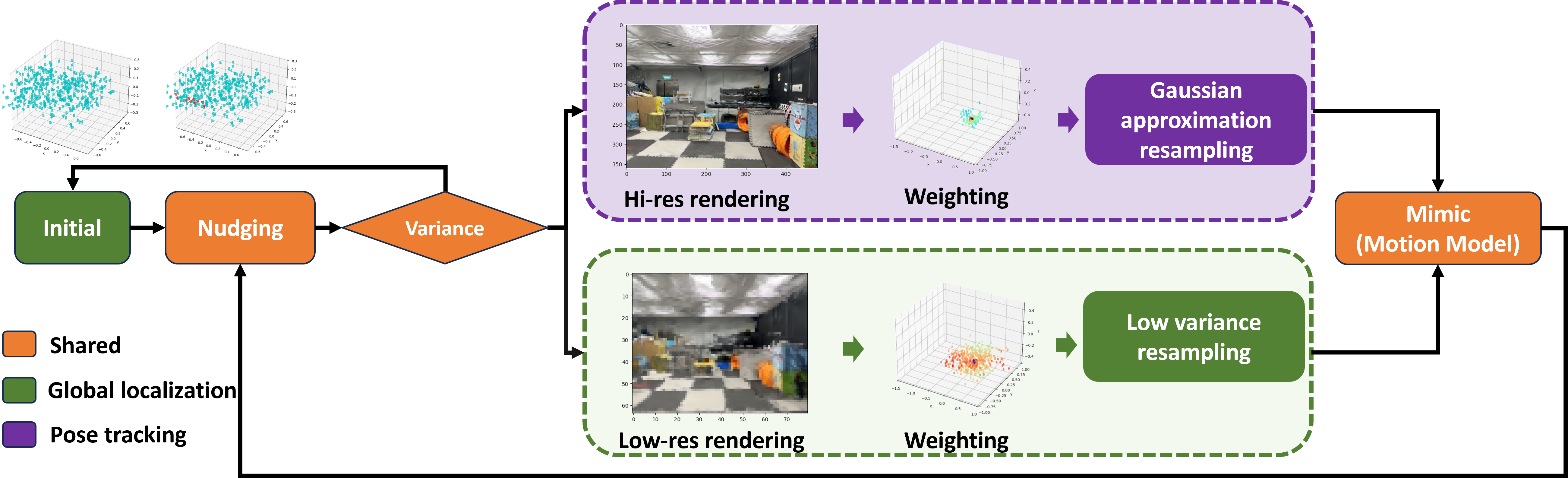}
  \caption{The pipeline of our designed NuRF framework. We first use radiance fields to generate images on anchor poses, store these images in a database, and vectorize them for retrieval. Then, a particle filter is built for robot localization in radiance fields. The nudging step uses retrieved results (from VPR) to guide the particles toward more confident states. The measurement model updates particle weights based on images that are observed and rendered by particles. The motion model adjusts particles according to robot motion. Our adaptive workflow enables switching between global localization and pose tracking to address the robot kidnapping problem.} 
  \label{workflow}
\end{figure*}

\section{Methodology} 
\label{sec:meth}

To address the outlined problems, we introduce the NuRF framework as depicted in Figure \ref{workflow}. This framework integrates anchor points in close proximity to the target image $I_\tau$ within the feature space during the resampling phase. By implementing this strategy, both Problem \ref{prob:nudge} and Problem \ref{prob:localization} are effectively addressed for real-world deployment. Additionally, the utilization of nudging particles facilitates the monitoring of particle dispersion, i.e., the variance of particles, which is used to switch the global localization and poser tracking in the radiance fields. 

\subsection{Radiance Fields-enhanced VPR Anchors}
\begin{figure*}[t]
    \centering
    \includegraphics[width=18cm]{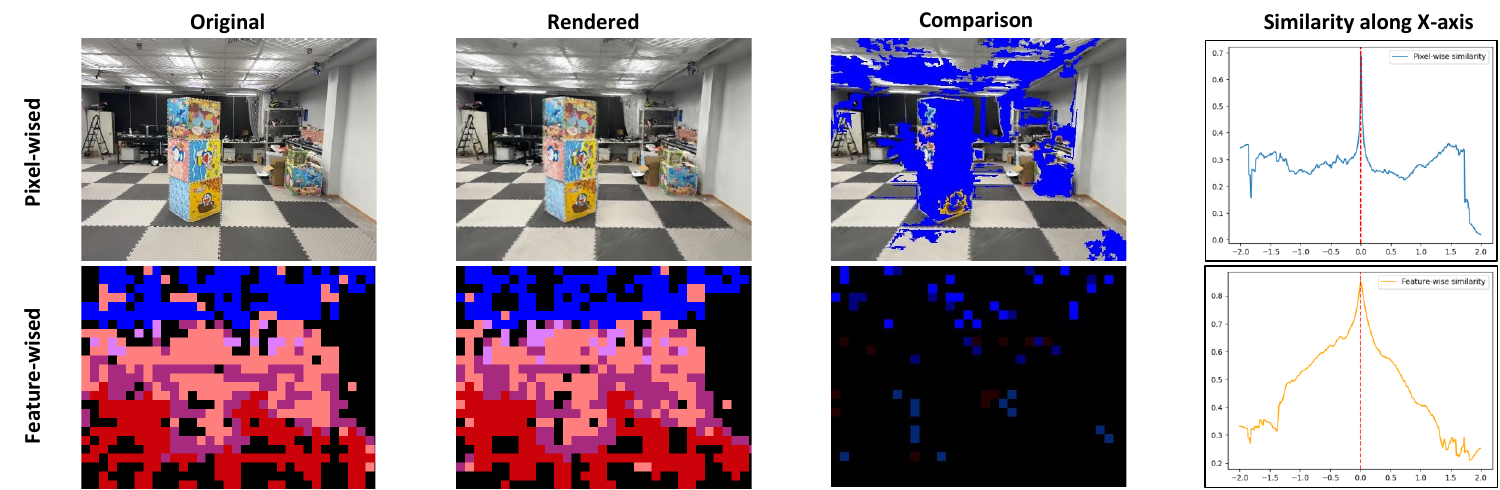}
    \caption{We show the original image and the rendered image in pixel space and feature space correspondingly. In the comparison column, the difference between the original image and the rendered image is highlighted in blue. We assess the similarity between a query image and images rendered in the same orientation but shifted along the X-axis using pixel-wise similarity and feature-wise similarity.}
    \label{dxyz}
\end{figure*}

Pixel-wised similarity is sensitive to translational movement, as shown in Figure~\ref{dxyz}. Even if the orientation error is $0$, only images rendered in a very small spatial neighborhood of the target image location can obtain a high similarity response. In global localization, when the number of particles is limited, the probability of randomly generated particles falling into this neighborhood is also very small, and the particle filter might fail to converge to the ground truth pose. 

\begin{figure}[t]
    \centering
    \includegraphics[width=8cm]{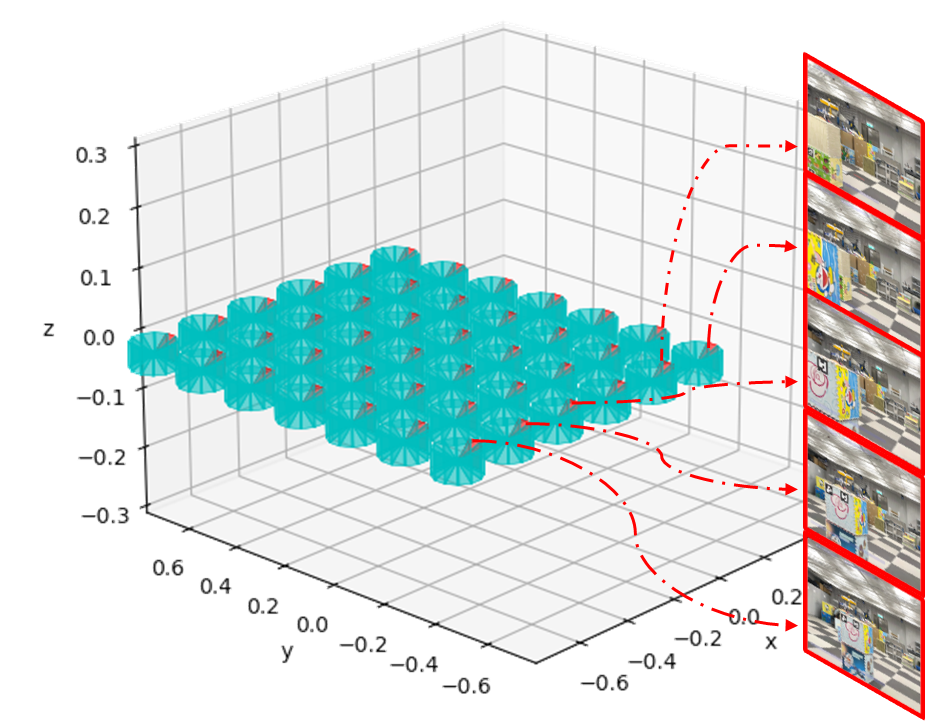}
    \caption{We display 504 anchor poses in 2D sub-manifold $S$, along with images rendered at these specific anchors for visualization. }
    \label{fig:aug}
\end{figure}

In order to overcome these problems and build an efficient NuRF (Problem \ref{prob:localization}), we introduce the VPR on anchor poses. As shown in Figure~\ref{fig:aug}, VPR technology requires a set of anchor points in the space and stores the images generated at these anchor points. Specifically, we detail the anchor setting in Algorithm \ref{alg:index}, the process initiates by sampling a set of 3-DoF anchors, denoted as $s = (x, y, \omega)$, from a 2D sub-manifold of SE(3). Subsequently, each anchor $s$ is converted into a 6-DoF pose $\xi = (\eta_1, \eta_2)$ by incorporating zero values for the z-axis, pitch, and roll angles. This conversion employs the exponential map for $\xi$ to SE(3), as follows:
\begin{equation}
        \exp{(\xi^{\wedge})} = \sum_{n = 0}^{\infty} \frac{1}{n!}(\xi^{\wedge})^n 
        = \begin{bmatrix}
            \boldsymbol{R} & \boldsymbol{t} \\
            \boldsymbol{0} & 1
        \end{bmatrix}
    \label{eq:exp}
\end{equation}
Following this, the rendered image $I$ is embedded into a feature space $\boldsymbol{F}$ using a pre-trained embedding function \cite{oquab2023dinov2}. Both the image feature $\boldsymbol{F}$ and the pose $T$ are subsequently stored in a database $\boldsymbol{D}$. 

\begin{algorithm}[t]
    \SetAlgoLined
    \SetKwInOut{Input}{Input}
    \SetKwInOut{Output}{Output}
    \SetKwFunction{Embedding}{Encode}
    
    \Input{$\mathcal{G},L,K,N,S$}
    \Output{$\boldsymbol{D}$}
    \BlankLine
    
    $\boldsymbol{T} \gets \emptyset$\;
    $\boldsymbol{F} \gets \emptyset$\;
    $\boldsymbol{S} \gets [(x_1,y_1,\omega_1),\dots,(x_N,y_N,\omega_N)]$\;
    
    \For{$s \in \boldsymbol{S}$}{
        $\xi \gets [s[1],s[2],0,s[3],0,0]$\;
        $T \gets \exp{(\xi^{\wedge})}$ \tcp*{Equation \eqref{eq:exp}}
        $I \gets L(T)$ \tcp*{Equation \eqref{eq:rendering}}
        $\boldsymbol{F} \gets \boldsymbol{F} \cup$ \Embedding{$I$}\;
        $\boldsymbol{T} \gets \boldsymbol{T} \cup T$\;
    }
    
    $\boldsymbol{D} \gets \{\boldsymbol{F}:\boldsymbol{T}\}$\;
    
    \Return{$\boldsymbol{D}$}\;
    
    \caption{Anchor Setting in Radiance Fields}
    \label{alg:index}
\end{algorithm}

\subsection{Nudging by Visual Place Recognition}
After generating the VPR anchors, we need to incorporate the reference information provided by the VPR results into the pixel-wised weighting particle filter to solve Problem \ref{prob:nudge}.
We adopt the definition of the nudging step provided by Akyildiz\cite{akyildiz2020nudging}, illustrated in Definition \ref{def:nudging}. In this definition, $x_{\tau} \in \mathsf{X}$ represents the system state at time $\tau$, and $y_{\tau}$ denotes the observations at time $\tau$. The state after the nudging step is denoted as $x^+$.
\begin{definition}
    A nudging operator $\alpha_{\tau}^{y_{\tau}}: \mathsf{X} \rightarrow  \mathsf{X}$ associated with the likelihood function $g_{\tau}(x)$ is a map such that if $x^+ =  \alpha_{\tau}^{y_{\tau}}(x)$, then $g_{\tau}(x^+) \geq g_{\tau}(x)$.
    \label{def:nudging}
\end{definition}

Intuitively, the nudging step adapts the generation of particles in a manner that is not compensated by the importance of weights but enhances the likelihood of accurate estimation. To achieve this, we have developed a novel technique termed VPR nudging.

\begin{figure*}[t]
    \centering
    \includegraphics[width=18cm]{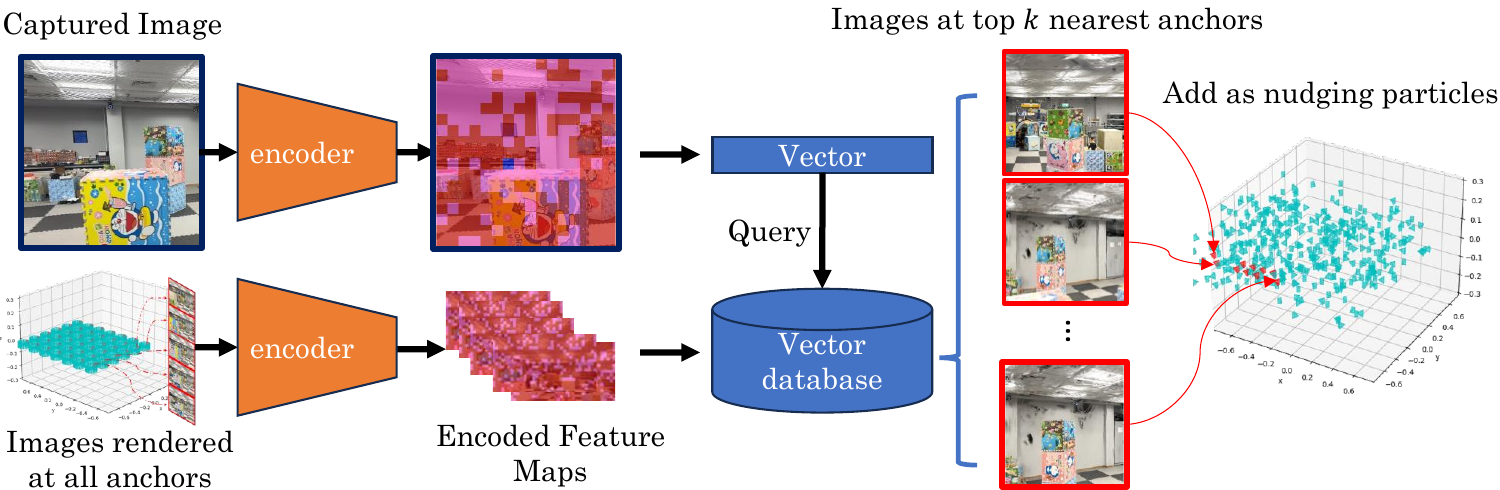}
    \caption{Pipeline of nudging particles using the feature information from VPR. Upon the observation of a new image, the ViT encoder transforms it into a feature map. This map is then vectorized and subjected to cosine distance calculations with the vectors stored in the VPR database. The camera pose corresponding to the $k$ vectors with the shortest cosine distance are retrieved and add into the particle set.} 
    \label{fig:nudge-pip}
\end{figure*}

Figure \ref{fig:nudge-pip} demonstrates the VPR nudging step, as illustrated in Definition \ref{def:nudging}. The algorithm inputs include the reference image database $\boldsymbol{D}$ at anchors, the observed image $I$, the number of nudging particles $M$, the radiance field $[\mathcal{G},L]$, and the low-resolution camera intrinsic $K^-$. The objective is to identify the top $M$ images in the database $D$ that are most similar to $I$ and to render low-resolution images $I_m$ at each corresponding pose $T_m$. 
If the weight of nudged particles $w_m$ exceeds the current average weight $\bar{w}$, the pose $T_m$ is incorporated into the nudging particle set $\varXi^+$.

The VPR nudging step modifies the stochastic generation of particles in a manner that is not offset by the importance weights. This strategic alteration enables the integration of valuable feature information without necessitating computationally intensive embedding operations on every image rendered by particles. This approach enhances both the accuracy and efficiency of the global localization.

\subsection{Adaptive Scheme in NuRF}

We introduce an adaptive scheme within the particle filter framework to switch the global localization and pose tracking. Initially, in global localization, we lower the resolution of both the input and rendered images. This reduces localization accuracy slightly but expands the area where high similarity responses are detected, lessening sensitivity to small movements. As the process iterates and the particles begin to converge, we maintain the original input image quality and enhance the resolution of the rendered images to improve pose tracking accuracy. This adaptive strategy dynamically adjusts the resolution based on the discrepancy between nudged and original particles. This approach effectively addresses the issues of particle deficiency in global localization and aids in automatic recovery from kidnapping scenarios.

Specifically, after each nudging step, we calculate the weighted variance of the current particles. This variance indicates how close the particle filter is to converging and reflects the diversity of particle distribution. It directly influences how we adjust the resolution in the image rendering process to optimize accuracy and performance. The weight variance is defined as follows:
\begin{align}
    \bar{T} &= \frac{1}{N} \sum_{i=1}^{N} w_iT_i \label{eq:weighted_mean}\\
    \Sigma_T &= \frac{1}{N}  \frac{\sum_{i=1}^{N} w_i(T_i - \bar{T})(T_i - \bar{T})^T}{\sum_{i=1}^{N} w_i} \label{eq:weighted_variance}
\end{align}
in which $\Sigma_T$ is a metric for assessing the convergence of the particle filter. When the weighted variance of the particles drops below a set threshold $\lambda$, the system shifts to a higher resolution setting ($K^+$), improving the accuracy of pose estimation. If the variance exceeds this threshold, the algorithm switches to a lower resolution setting ($K^-$), reducing computational load and increasing the efficiency of robot localization. The strategy is described as:
\begin{equation}
    K = \left\{ 
    \begin{array}{r c l}
    K^+ & & {\Sigma_T \leq \lambda} \\
    K^- & & {\Sigma_T >\lambda} \\
    \end{array}
    \right.
\end{equation}

The system pipeline, as depicted in Figure \ref{workflow}, seamlessly integrates VPR nudging and adaptive resolution rendering techniques. Particle adjustments span extensive areas using low-resolution rendering for the global localization phase at a lower resolution ($K=K^-$). The resampling weights for this phase are calculated as follows:
\begin{equation}
w_{\tau,j} = \frac{\text{LOSS}(I_{\tau}, L(T_{\tau,j}))}{\sum_{i=1}^{N} \text{LOSS}(I_{\tau}, L(T_{\tau,i}))}
\label{eq:global_weight}
\end{equation}

Conversely, during the local pose tracking phase at a higher resolution ($K=K^+$), the particle updates are refined in more confined areas using high-resolution rendering. The resampling weights in this phase are determined by:
\begin{equation}
w_{\tau,j} = \mathcal{N}(T_{\tau,j}; \mu_{\tau}, \Sigma_{\tau})
\label{eq:local_weight}
\end{equation}
where $\mu_{\tau}$ and $\Sigma_{\tau}$ represent the weighted mean and variance, respectively, detailed in Equations \eqref{eq:weighted_mean} and \eqref{eq:weighted_variance}. The Gaussian approximate resampling technique, often used in this phase, is particularly suited for environments expected to exhibit a single-peak distribution. This method is crucial in high-resolution settings, where even minor distortions such as rotations and translations can significantly impact particle convergence. The strategies for both global localization and local pose tracking, thoroughly described in Algorithm \ref{alg:global} and Algorithm \ref{alg:local}, ensure robust performance across varying environmental complexities.

\begin{algorithm}[t]
    \SetAlgoLined
    \SetKwInOut{Input}{Input}
    \SetKwFunction{SSIM}{SSIM}
    \SetKwFunction{Resampling}{resampling}
    \SetKwFunction{Alpha}{alpha}
    
    \Input{$\varXi_{\tau-1},I_{\tau},\mathcal{G},L,D,M$}
    \BlankLine
    
    \For{$\zeta_{\tau-1,i} \in \varXi_{\tau-1}$}{
        $T_{\tau,i} \gets T_{\tau-1} H_{\tau}$ \tcp*{Equation \eqref{eq:pose}}
        $\hat{I}_{\tau,i}\gets L(T_{\tau,i})$ \tcp*{Equation \eqref{eq:rendering}}
        $w_{\tau,i} \gets w_{\tau,i-1}$  \tcp*{Equation \eqref{eq:likeli}}
    }
    
    $\varXi^+_{\tau} \gets \alpha(D,I_{\tau},M,\mathcal{G},L,K^-)$ \tcp*{Nudging}
    
    $\varXi^\dagger_{\tau} \gets \emptyset$
    
    \For{$\zeta_{\tau,i} \in \varXi_{\tau}$}{
        $\zeta^\dagger_{\tau,i} \gets $ \Resampling{$f(z;\varXi^+)$} 
        
        $\varXi^\dagger_{\tau,i} \gets \zeta^\dagger_{\tau,i}$
    }
    
    \Return{$\varXi^\dagger_{\tau}$}\;
    
    \caption{Global Localization}
    \label{alg:global}
\end{algorithm}

\begin{algorithm}[t]
    \SetAlgoLined
    \SetKwInOut{Input}{Input}
    \SetKwFunction{SSIM}{SSIM}
    \SetKwFunction{Resampling}{resampling}
    \SetKwFunction{Alpha}{alpha}
    \SetKwFunction{WeightedMean}{E}
    \SetKwFunction{WeightedVariance}{Var}
    
    \Input{$\varXi_{\tau-1},I_{\tau},\mathcal{G},L,D,M$}
    \BlankLine
    
    \For{$\zeta_{\tau-1,i} \in \varXi_{\tau-1}$}{
        $T_{\tau,i} \gets T_{\tau-1} H_{\tau}$ \tcp*{Equation \eqref{eq:pose}}
        $\hat{I}_{\tau,i}\gets L(T_{\tau,i})$ \tcp*{Equation \eqref{eq:rendering}}
        $w_{\tau,i} \gets w_{\tau,i-1}$  \tcp*{Equation \eqref{eq:likeli}}
    }
    
    $\varXi^+_{\tau} \gets \alpha(D,I_{\tau},M,\mathcal{G},L,K^+)$ \tcp*{Nudging}
    
    $\mu_{\tau} \gets$ \WeightedMean{$\varXi^+_{\tau}$} \tcp*{Equation \eqref{eq:weighted_mean}}
    
    $\Sigma_{\tau} \gets$ \WeightedVariance{$\varXi^+_{\tau}$} \tcp*{Equation \eqref{eq:weighted_variance}}
    
    $\varXi^\dagger_{\tau} \gets \emptyset$
    
    \For{$\zeta_{\tau,i} \in \varXi_{\tau}$}{
        $\zeta^\dagger_{\tau,i} \gets $ \Resampling{$\mathcal{N}(\mu_{\tau},\Sigma_{\tau})$} 

        $\varXi^\dagger_{\tau,i} \gets \zeta^\dagger_{\tau,i}$
    }
    
    \Return{$\varXi^\dagger_{\tau}$}\;
    
    \caption{Pose Tracking}
    \label{alg:local}
\end{algorithm}



\section{Experiments and Evaluation}
\label{sec:exp}

We validate the proposed NuRF in a motion capture room, which is a typical indoor environment with cluttered objects. The experiments are designed to assess the performance in terms of global localization and pose tracking, i.e., the convergence from scratch and local tracking accuracy. 
All the experiments are conducted on a blimp robot \cite{tao2021swing}.


\subsection{Experimental Setup}

\subsubsection{Radiance Field in Indoor Environments}
\begin{figure}[t]
    \centering
    \subfigure[Experimental setup]{\includegraphics[width=8cm]{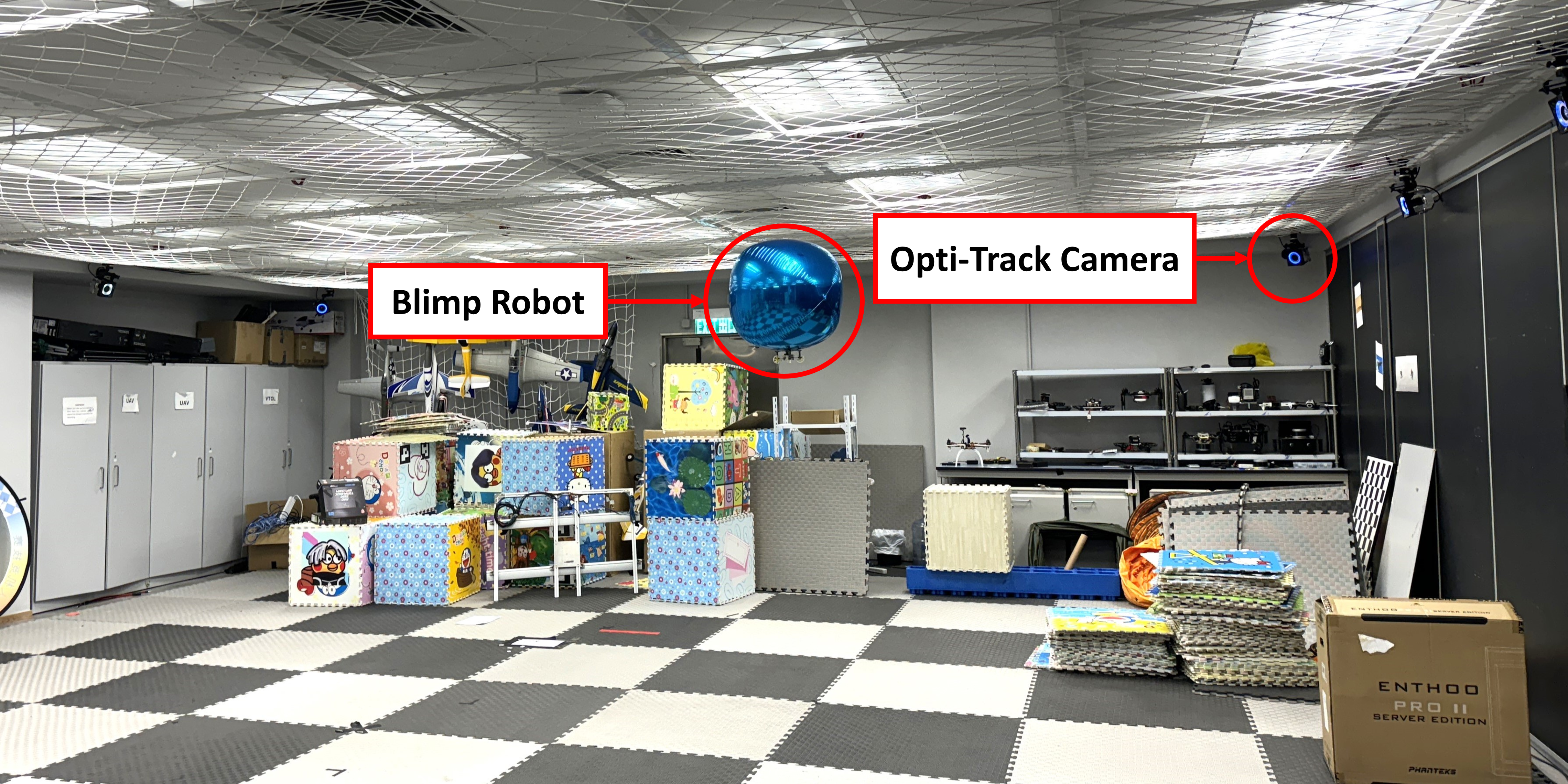}\label{arena}} 
    
    \subfigure[Rendered image]{\includegraphics[width=8cm]{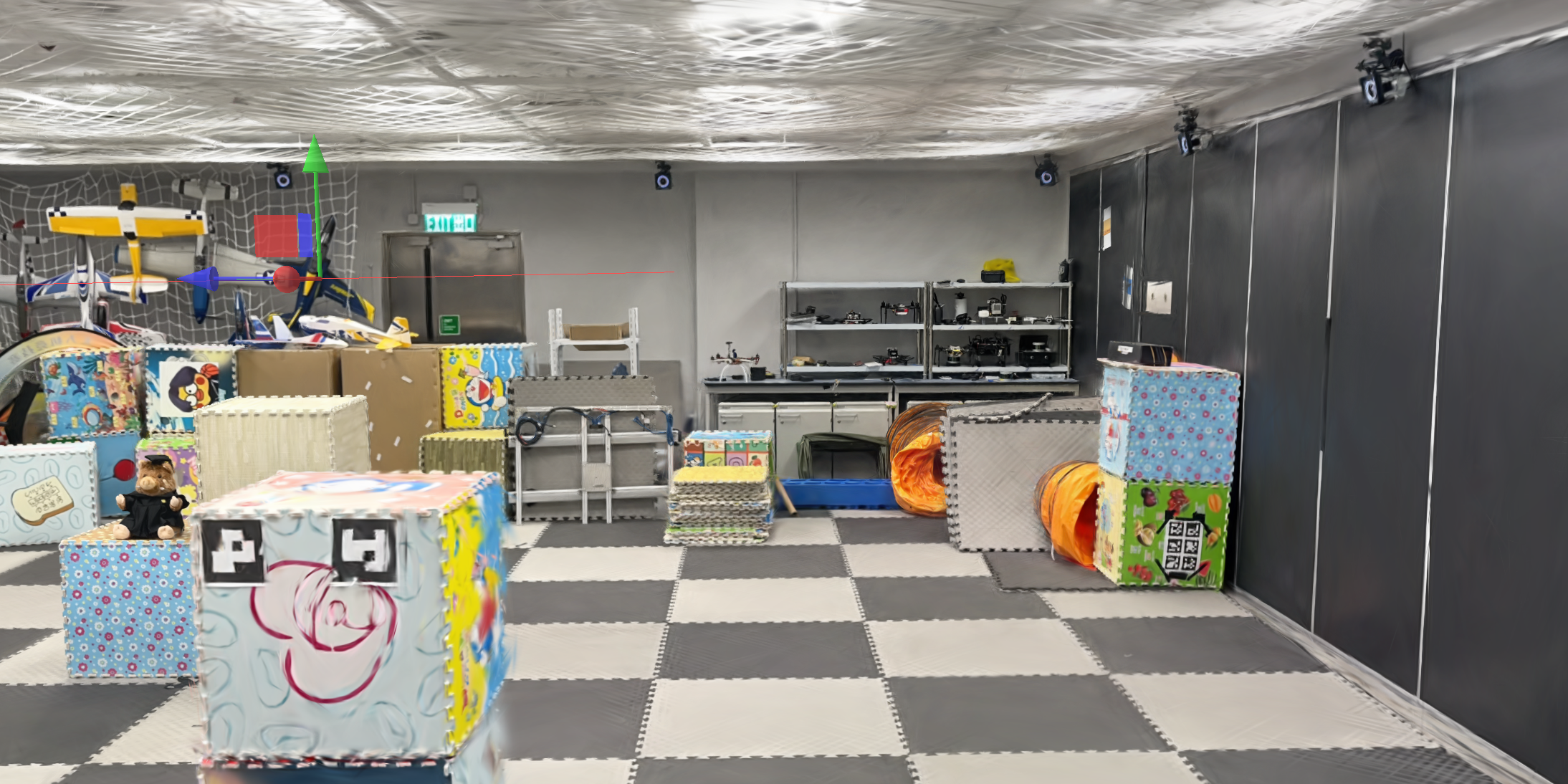}\label{GS_model}} 
    
    \caption{(a) A Blimp robot in our motion capture room for experimental evaluation. (b) A rendered image in the motion capture room that closely mimics the real-world scenario. }
\end{figure}
Our method relies on a pre-trained 3D radiance field as a pre-built map for robot localization. We directly utilize the 3DGS~\cite{kerbl20233d}, a real-time radiance field rendering technique known for its state-of-the-art performance. The input of the 3DGS is a trajectory consisting of continuous camera poses and corresponding images at each pose, which can be easily obtained on our blimp robot.

Our blimp robot is equipped with one camera for monocular localization. The robot was operated remotely within a 4-meter by 5-meter fly arena, as shown in Figure \ref{arena}. A motion capture system is installed. to obtain the ground truth poses. To generate the 3DGS of the room, 534 images with a $1440 \times 1920$ resolution are captured, and the rendered image is shown in Figure \ref{GS_model}.


\begin{figure*}[t]
  \centering
  \includegraphics[width=16cm]{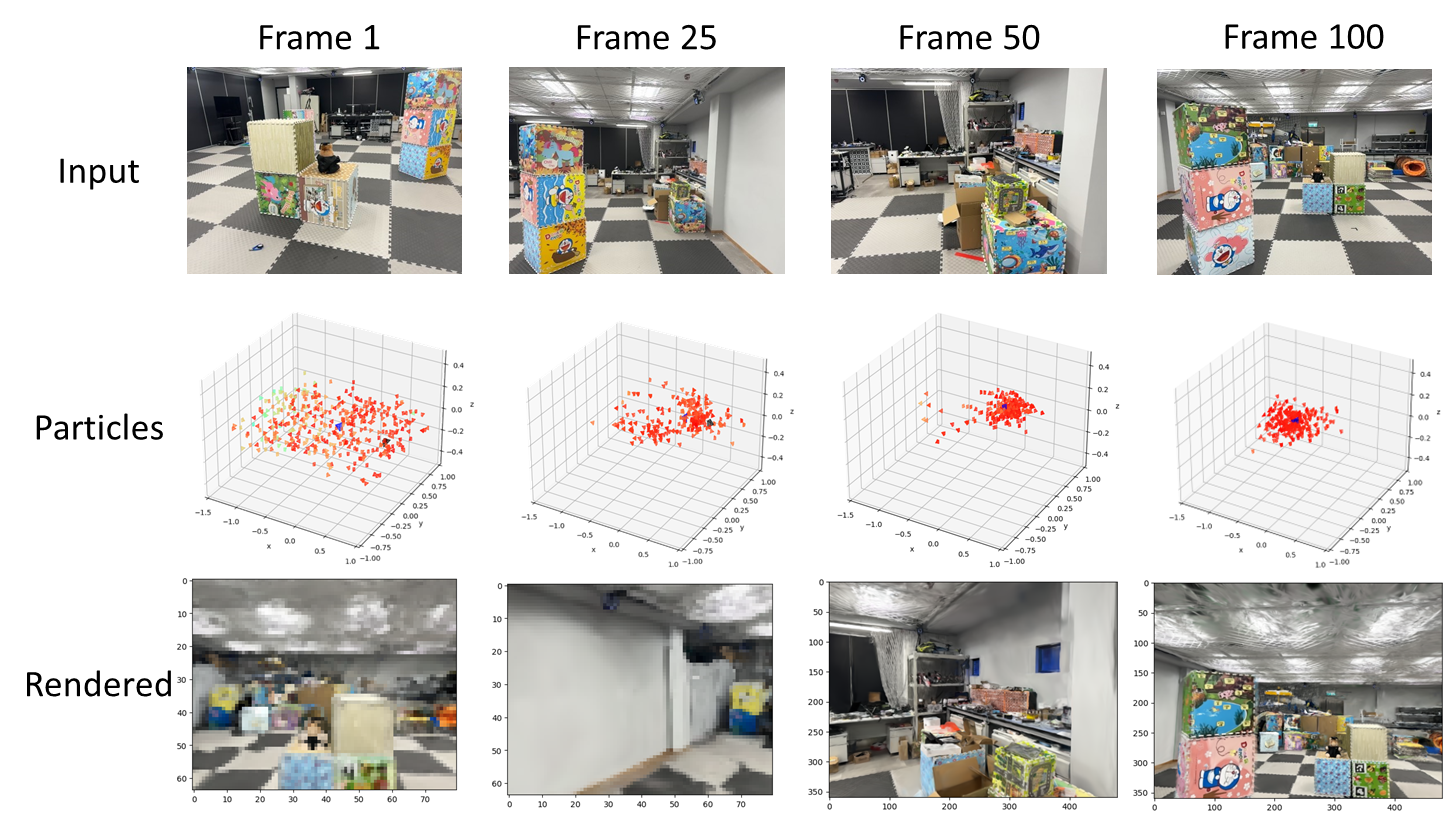}
  \caption{A case study for global visual localization in radiance fields. Top: Sequential observed images of the robot at frame 1, frame 25, frame 50, and frame 100. Middle: Particle filter states for each frame are shown. The larger blue pyramid indicates the estimated pose, while the black pyramid represents the ground truth pose. Smaller pyramids symbolize the particles, colored according to the visible spectrum, with red indicating the highest similarity and purple the lowest. Bottom: Rendered images at the estimated pose are shown, with the rendering resolution adapting from low to high based on the variance of particles.}
  \label{global_g2}
\end{figure*}


\subsubsection{{Baseline}}
We compare NuRF with Loc-NeRF~\cite{maggio2023loc} in both global visual localization and pose tracking tasks to demonstrate its superior accuracy and speed. NuRF utilizes 3DGS as the map, which renders much faster than NeRF~\cite{kerbl20233d} in the original Loc-NeRF. Therefore, to ensure a fair comparison, we replace the NeRF with the 3DGS map in the Loc-NeRF system.

\subsubsection{Implementation Details}

The experiments are conducted with Pytorch-based implementation. The computing hardware comprises an Intel CPU i7-13700KF and a GeForce RTX4080 GPU. The input image size was set to $680 \times 800$, with a render resolution of $64 \times 80$ for global localization and a render resolution of $680 \times 800$ for pose tracking. The thresholds for global localization and pose tracking switching were set to $\lambda^+=5$.

\subsection{Global Visual Localization}\label{sec:exp1}

To evaluate the NuRF method's performance in global localization, we conducted experiments with 20 randomly chosen sets of image sequences. Each set contained 100 frames depicting continuous motion.

A demonstration for global visual localization in radiance fields with NuRF is shown in Figure \ref{global_g2}, the first column illustrates the utilization of 400 particles. The initial positions of these particles were obtained by uniformly perturbing the dimensions of the room, which has a height of $1.8m$, a length of $5m$, and a width of $4m$. For orientation, the yaw angles were uniformly initialized within the range of $[-180^{\circ},180^{\circ}]$, while the pitch and roll angles for all particles were set to $0^{\circ}$. Once the experiment begins, the motion model provides a 6-DoF (six degrees of freedom) motion matrix to update the pose of the particle swarm (depicted in the second to fourth columns of Figure \ref{global_g2}). The resampling process also occurs in the 6-DoF space, meaning that the algorithm optimizes the particles in a complete 6-DoF state and performs only one update per image.
\begin{figure}[t]
    \centering
    \includegraphics[width=8cm]{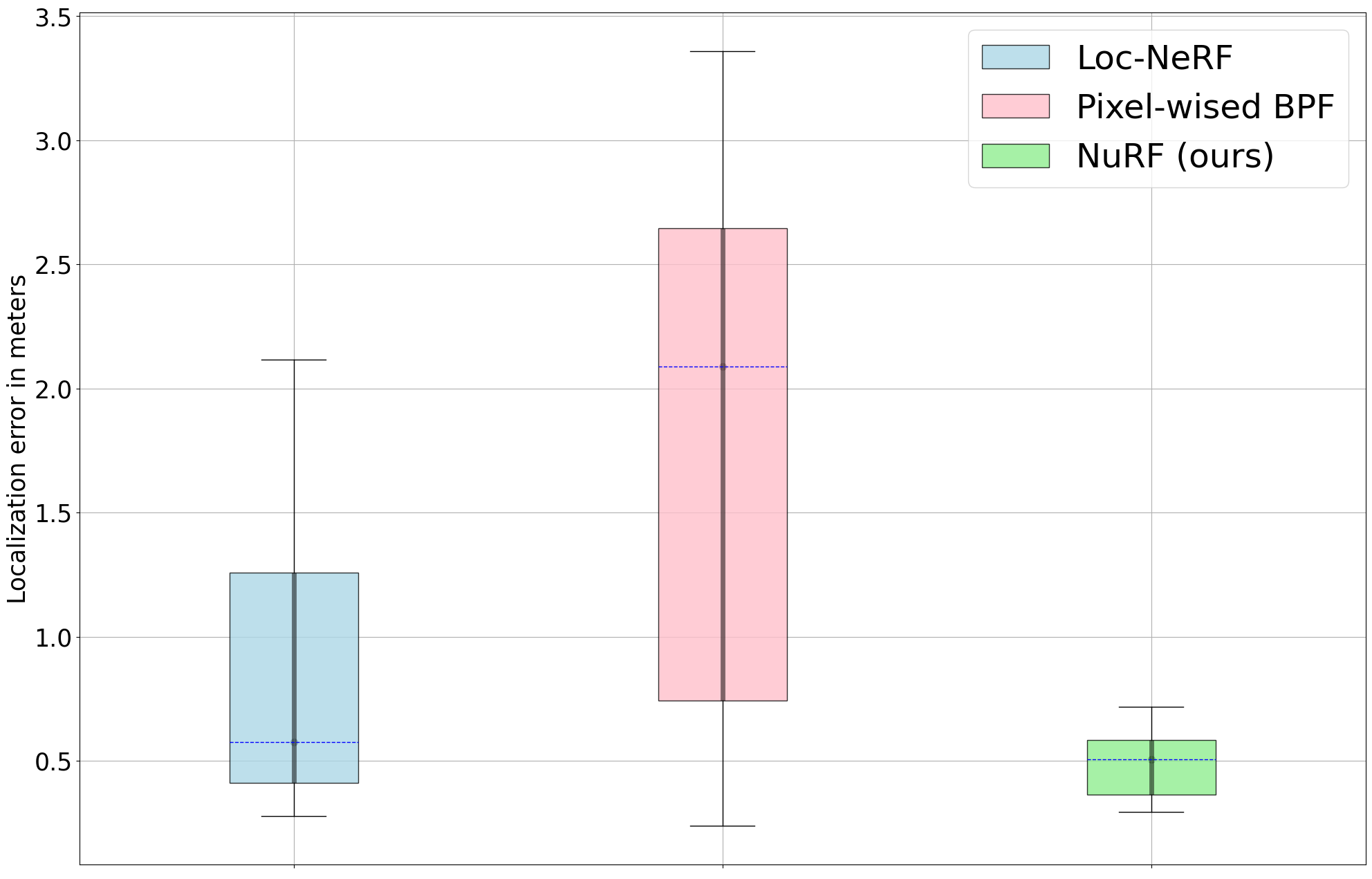}
    \caption{Average localization errors of NuRF and comparisons over 20 trials.}
    \label{fig:msg}
\end{figure}
\begin{figure}[t]
    \centering
    \includegraphics[width=8cm]{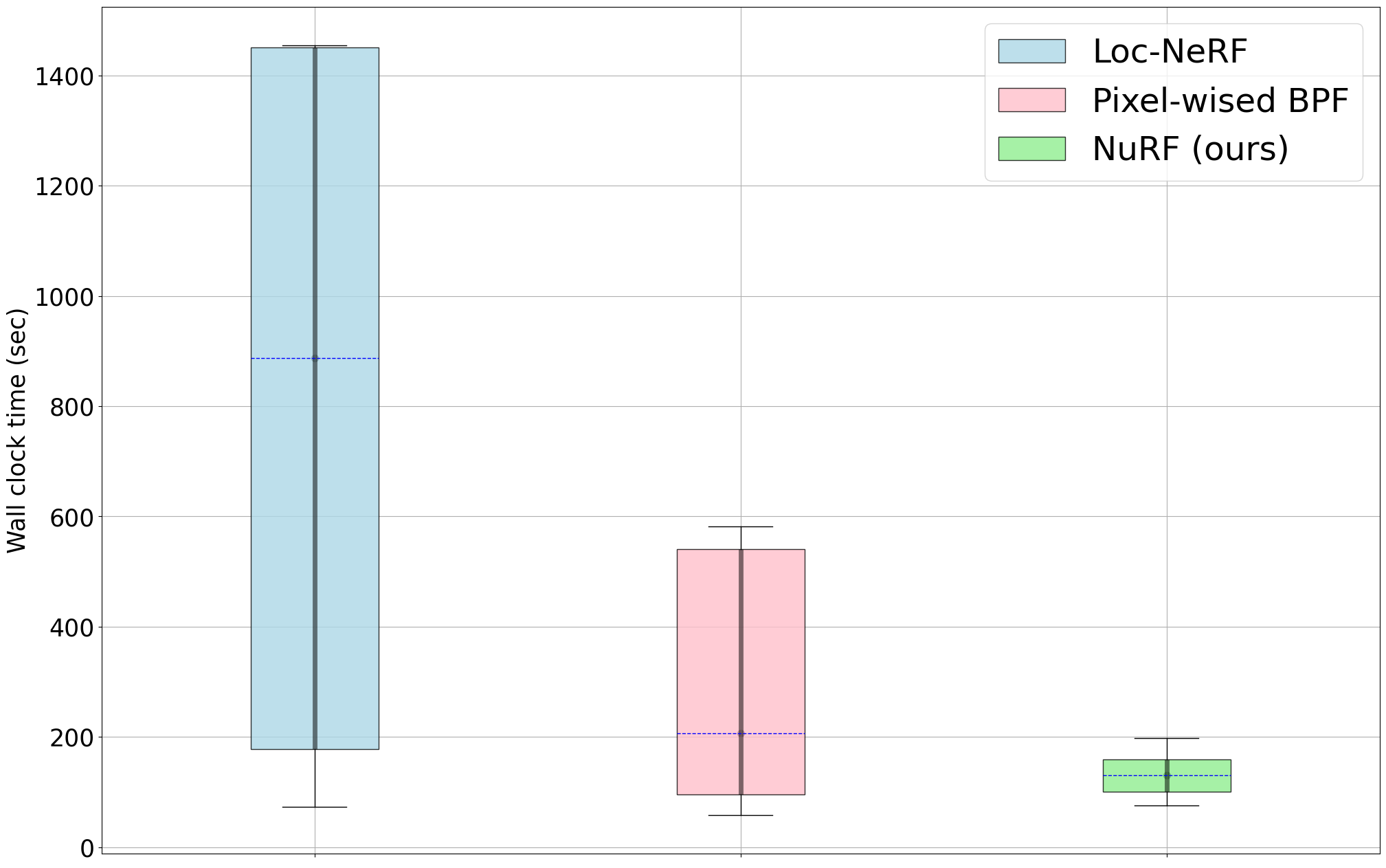}
    \caption{Average time costs of NuRF and comparisons over 20 trials.}
    \label{fig:wct}
\end{figure}

We conducted a test to compare the global localization capabilities of the Loc-NeRF algorithm with ours using the same image sequence. Additionally, to validate the modules in NuRF, we performed ablation experiments by omitting the nudging step and utilizing only the bootstrap particle filter (BPF) with pixel-wise weights for global localization performance. The experimental results are displayed in Figures~\ref{fig:msg} and \ref{fig:wct}. In the box plot, the upper and lower edges indicate the maximum and minimum localization errors from the 20 experiments, while the box length represents the interquartile range. The median is denoted by a blue dashed line in the center of the box. Figure~\ref{fig:msg} shows that the median localization error for NuRF is approximately 0.5 m, which is lower than that of Loc-NeRF and the BPF when the VPR nudging step is removed. Additionally, NuRF exhibits the most stable localization error across the 20 experiments, with significantly lower variance compared to the other two algorithms. Figure~\ref{fig:aug} further highlights the operational efficiency of NuRF, illustrating that its convergence time is notably shorter than that of Loc-NeRF and the standard BPF without the nudging step.

\subsection{Pose Tracking in Radiance Fields}

\begin{figure}[t]
    \centering
    \includegraphics[width=8.5cm]{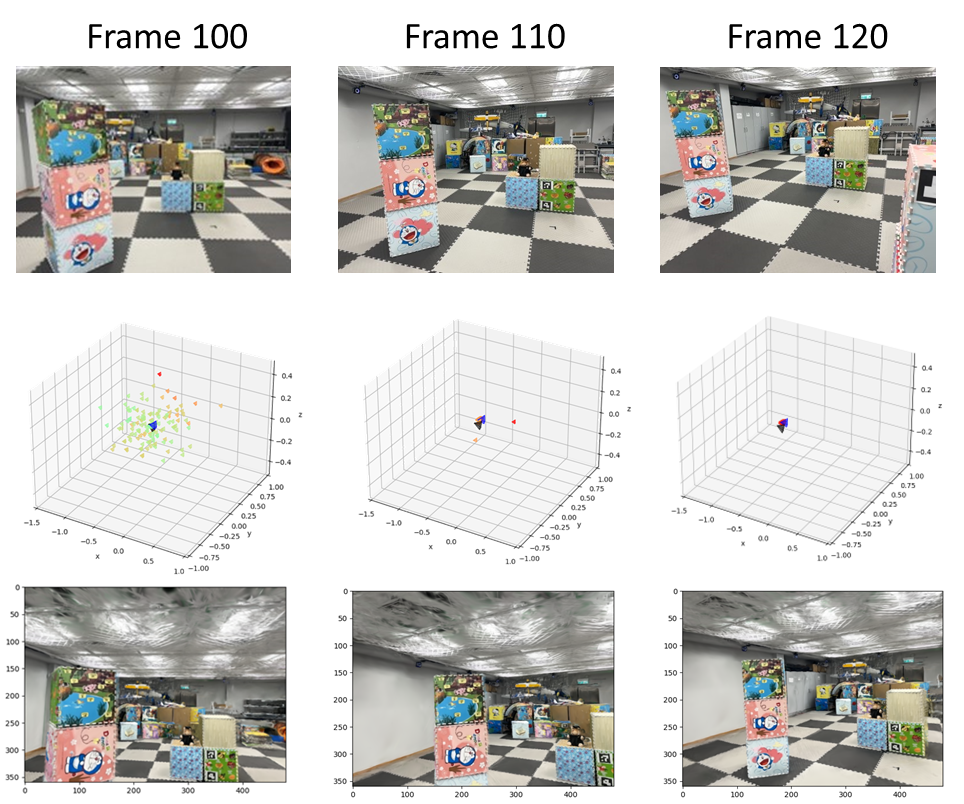}
    \caption{A case study of pose tracking. Top: observed images of the robot at key frames: frame 100, frame 110, and frame 120. Middle: The particle filter states at each frame, following the same color settings in Figure~\ref{global_g2}.  Bottom: the rendered images at the estimated poses, with changes in rendering resolution based on the variance of the particles.}
    \label{tracking_g2}
\end{figure}

In this section, we evaluate the performance of the NuRF method in a pose tracking task using images captured from a blimp. We initiate the tracking process with the pose estimation obtained in Section~\ref{sec:exp1}. We employ $200$ particles for the pose tracking task and initialize their positions by sampling from a normal distribution centered around the estimated position. The variances for the position dimensions are set to $[0.2, 0.2, 0.1]$ respectively. Similarly, we sample the yaw, pitch, and roll angles from normal distributions centered around the estimated rotation. The variances for the angles are set to $[5, 1, 1]$ respectively. Figure \ref{tracking_g2} illustrates a case study of tracking process, where only one update is performed per image.

The demonstration presented in the first column of Figure \ref{tracking_g2} shows the re-generation of the particles around the initially estimated position when transitioning from global localization to positional tracking. In the second column of Figure \ref{tracking_g2}, we observe the convergence of the particles after running the NuRF algorithm for 10 frames. In the pose tracking phase, the VPR system continues to generate particles. However, these particles are solely utilized to detect potential abduction issues and are not involved in the resampling process.

\begin{figure}[t]
    \centering
    \subfigure[Translation]{\includegraphics[width=8cm]{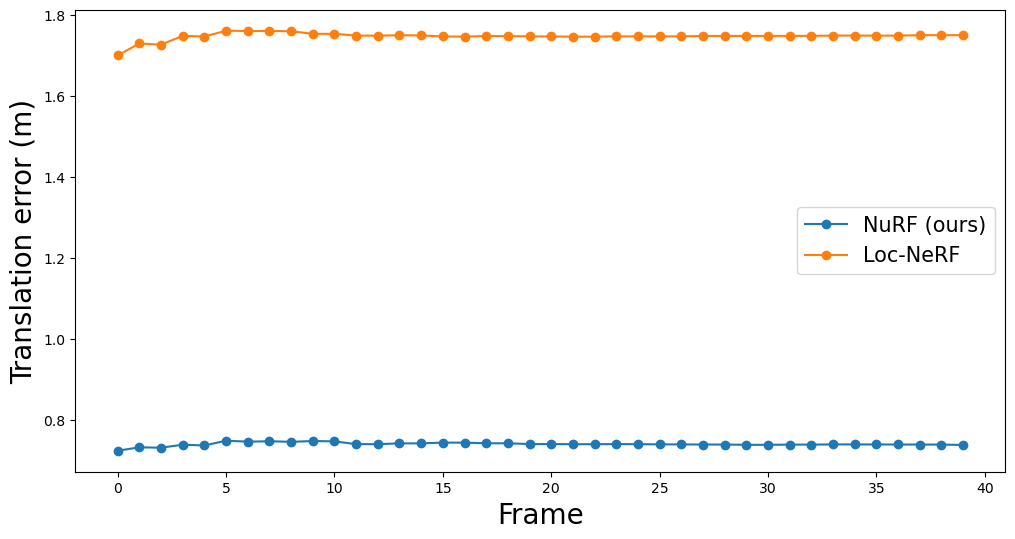}\label{tracking_t}}
    \subfigure[Roll]{\includegraphics[width=8cm]{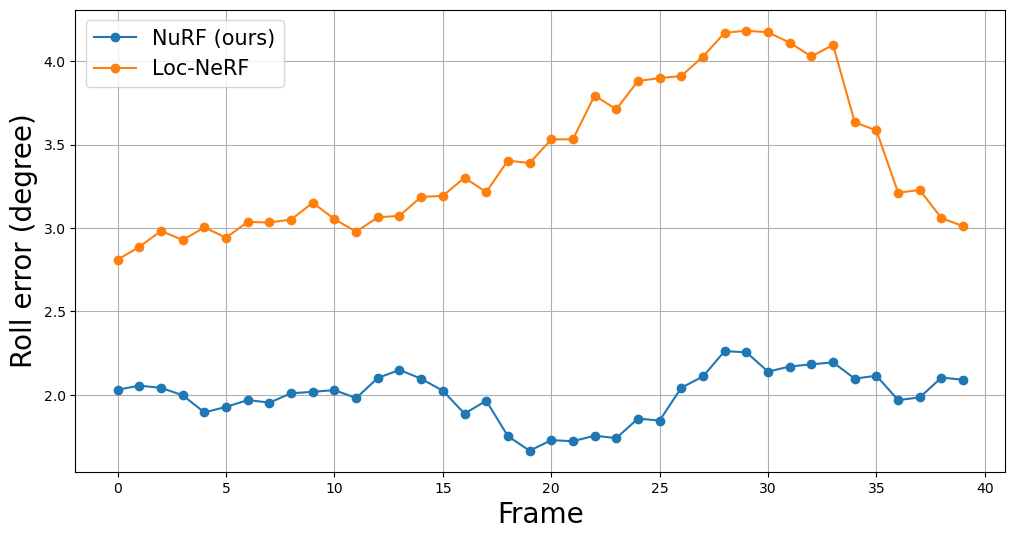}\label{tracking_r}}  
    \subfigure[Pitch]{\includegraphics[width=8cm]{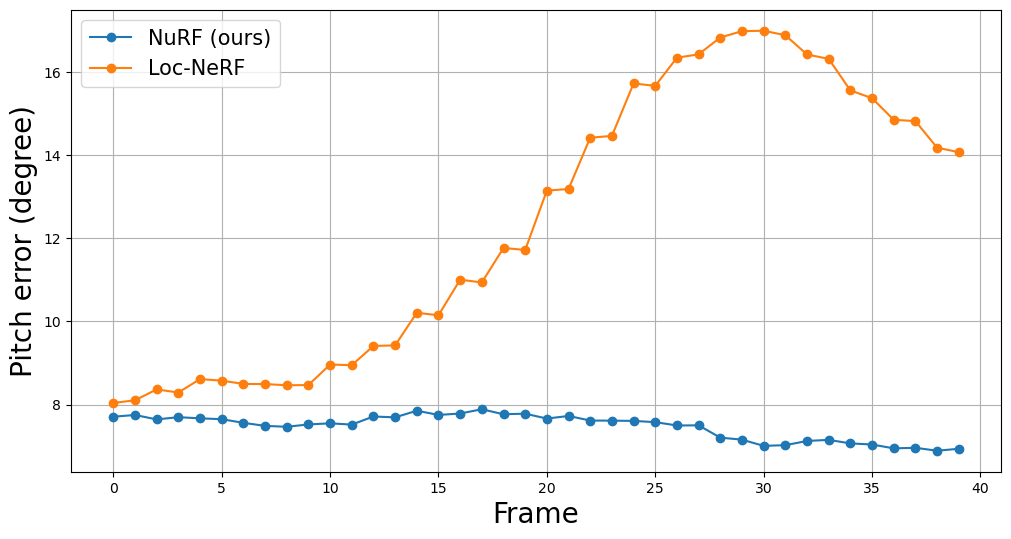}\label{tracking_p}} 
    \subfigure[Yaw]{\includegraphics[width=8cm]{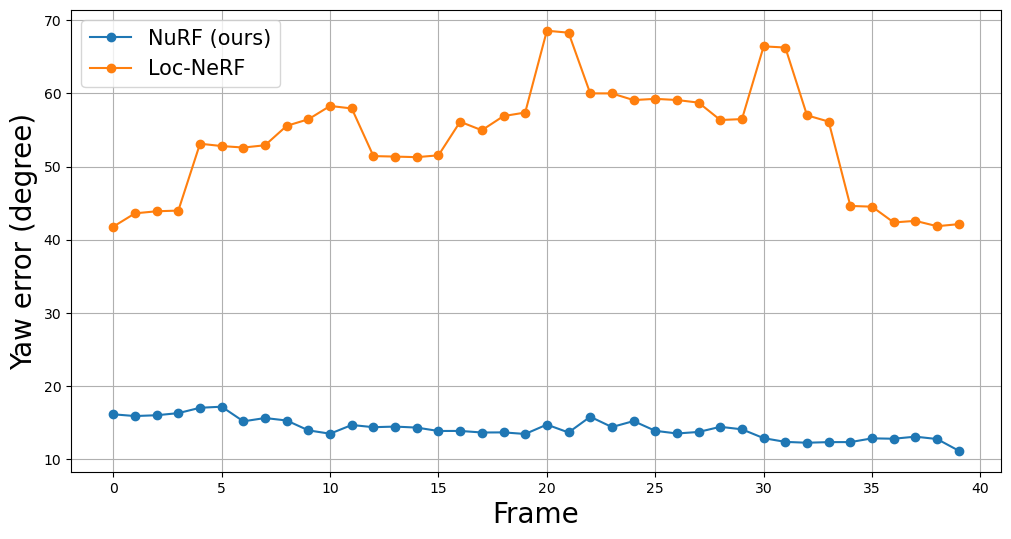}\label{tracking_y}} 
    \caption{Average translational and orientational error of the robot position over 40 frames of 20 trials.}
    \label{fig:tracking_o}
\end{figure}

To validate the effectiveness of the NuRF, we conduct experiments on 20 randomly selected trajectories consisting of 40 frames each and repeat the same experiment on Loc-NeRF and BPF with these data. The results of the experiment are presented in Table~\ref{tb:ab1}, which also shows the mean attitude tracking error and position error for each algorithm after 40 time steps. It can be seen in Figure~\ref{fig:tracking_o} that the algorithm NuRF has smaller tracking error than Loc-NeRF in pitch, yaw, and 3D spatial position.

\textcolor{blue}{
\begin{table}[t]
\caption{Average tracking error of various algorithms for 20 trials.}
\centering
\begin{tabular}{ccccc}
\hline
\hline
                & \multicolumn{3}{c}{\begin{tabular}[c]{@{}c@{}}Average orientation \\ tracking error (degree)\end{tabular}} & \multicolumn{1}{l}{\begin{tabular}[c]{@{}l@{}}Average position \\ tracking error (m)\end{tabular}} \\ \cline{2-4}
                & Roll                              & Pitch                             & Yaw                               &                                                                                                    \\ \hline
Loc-NeRF        & 3.51                     & 12.93                              & 55.91                              & 1.72                                                                                               \\
NuRF (ours)     & \textbf{2.17}                              & \textbf{7.81}                     & \textbf{12.18}                     & \textbf{0.64}                                                                                      \\ 
\hline
\hline
\end{tabular}
\label{tb:ab1}
\end{table}
}

\subsection{Ablated Experiments}
To validate the modules in NuRF, this section conducts ablated experiments on visual place recognition and explores the effect of the number of anchor points on global localization performance and the image resolution on global pose tracking performance. 

The ablated experiment is designed to evaluate global localization capabilities under two specific conditions: without VPR nudging (which is degenerated to pixel-wised BPF) and with attenuated VPR nudging (where the number of nudging particles was halved and the VPR anchor point count was reduced to 504). The results show in Table~\ref{tb:ab2} that if there are no anchored points, i.e., no nudging step is performed, there is a noticeable decline in the accuracy. Specifically, the translational mean square error of position estimation decreases from 0.642 $m$ to 1.204 $m$ when using NuRF for global localization. In terms of efficiency, the experimental results demonstrate that the nudging step leads to a slight increase in the time required for a single particle update, approximately 1 second. However, it significantly reduces the number of iteration steps needed for the convergence.
\textcolor{blue}{
\begin{table}[t]
\caption{Result of ablated experiment for 20 trials.}
\centering
\begin{tabular}{ccc}
\hline
\hline
                & \begin{tabular}[c]{@{}c@{}}Average position \\ tracking error (m)\end{tabular} & \begin{tabular}[c]{@{}c@{}}Average run \\ frequncy (Hz)\end{tabular} \\ \hline
Pixel-wised BPF & 1.204                                                                           & 0.0909                                                               \\
504-NuRF        & 0.79                                                                           & 0.0769                                                               \\
2502-NuRF       & 0.64                                                                  & 0.0706                                                               \\ 
\hline
\hline
\end{tabular}
\label{tb:ab2}
\end{table}
}


\section{Conclusions and  Future Study}
\label{sec:conclusions}

We present NuRF, a nudged particle filter designed for robot visual localization in radiance fields. Our approach incorporates key insights, including visual place recognition for nudging and an adaptive scheme for both global localization and pose tracking. To evaluate the effectiveness of the NuRF framework, we perform real-world experiments and provide comparisons and ablated studies. These experiments serve to validate the effectiveness of NuRF in indoor environments.

The proposed NuRF still has certain limitations for practical use. One key limitation is its deployment in texture-poor environments, where visual place recognition might fail due to the lack of sufficient visual features. Another limitation lies in the operational efficiency of NuRF. The main efficiency bottleneck is the rendering speed of images using NeRF. Currently, NuRF achieves a localization frequency of nearly 0.1 Hz. While this frequency is acceptable for low-speed ground mobile robots and indoor blimps, it poses challenges for integration into high-speed robotic platforms such as drones or self-driving cars.

We consider several promising directions to advance NuRF. One promising direction is the use of incremental 3D Gaussian Splatting~\cite{in3dgs}, which could help adapt and refine the NuRF system for larger-scale environments. The use of Gaussian Splatting could also improve the operational efficiency compared to the original NeRF. Another direction involves integrating planning capabilities within radiance fields to enable full navigation. Additionally, future research will focus on engineering efforts to achieve real-time robot localization, which is crucial for deploying NuRF in high-speed robotic platforms and dynamic environments.

\bibliographystyle{IEEEtran}
\bibliography{root}

\end{document}